\documentclass[conference]{IEEEtran}
\IEEEoverridecommandlockouts

\newif\ifarxiv
\arxivtrue

\usepackage{cite}
\usepackage{amsmath,amssymb,amsfonts}
\usepackage{algorithm}
\usepackage{algorithmic}
\usepackage{graphicx}
\usepackage[caption=false,font=footnotesize]{subfig}
\usepackage{textcomp}
\usepackage{xcolor}
\usepackage{url}

\renewcommand{\baselinestretch}{0.94}
\makeatletter
\renewcommand{\fs@ruled}{%
  \def\@fs@cfont{\bfseries}%
  \let\@fs@capt\floatc@ruled
  \def\@fs@pre{\kern 1.3mm\hrule height.8pt depth0pt \kern2pt}%
  \def\@fs@post{\kern2pt\hrule\relax}%
  \def\@fs@mid{\kern2pt\hrule\kern2pt}%
  \let\@fs@iftopcapt\iftrue
}

\def\BibTeX{{\rm B\kern-.05em{\sc i\kern-.025em b}\kern-.08em
    T\kern-.1667em\lower.7ex\hbox{E}\kern-.125emX}}

\ifarxiv
\usepackage[breaklinks,colorlinks]{hyperref}
\hypersetup{
  pdftitle={Error-Characteristic Analysis \\and Context-Aware Demodulation for Event-Camera-Based VLC},
  pdfsubject={Computer Vision, Robotics},
  pdfauthor={Ryota Soga, Tsukasa Shimizu, Shintaro Shiba, Quan Kong, Shan Lu, Takaya Yamazato},
  pdfkeywords={Event Cameras, Visible Light Communication, Optical Communication}
}
\usepackage[absolute]{textpos}
\else
\usepackage[pagebackref,breaklinks,colorlinks]{hyperref}
\fi

\begin{document}

\ifarxiv
\definecolor{somegray}{gray}{0.5}
\newcommand{\darkgrayed}[1]{\textcolor{somegray}{#1}}
\begin{textblock}{11.5}(2.25, 0.3)  
\begin{center}
\darkgrayed{This paper has been accepted for publication at the IEEE GLOBECOM 2026, \\
Macau S.A.R., China, 2026.
\copyright IEEE}
\end{center}
\end{textblock}
\fi

\title{Error-Characteristic Analysis \\and Context-Aware Demodulation
\\for Event-Camera-Based VLC}

\author{
\IEEEauthorblockN{
Ryota Soga\IEEEauthorrefmark{1},
Tsukasa Shimizu\IEEEauthorrefmark{2},
Shintaro Shiba\IEEEauthorrefmark{1},
Quan Kong\IEEEauthorrefmark{3},
Shan Lu\IEEEauthorrefmark{1},
Takaya Yamazato\IEEEauthorrefmark{1}
}
\IEEEauthorblockA{
\IEEEauthorrefmark{1}
Graduate School of Engineering, Nagoya University, Nagoya, Japan\\
\{soga, lu, yamazato\}@yamazato.nuee.nagoya-u.ac.jp\\
shiba@akg.t.u-tokyo.ac.jp}

\IEEEauthorblockA{
\IEEEauthorrefmark{2}
TOYOTA MOTOR CORPORATION, Toyota, Japan\\
tsukasa\_shimizu@mail.toyota.co.jp}
\IEEEauthorblockA{
\IEEEauthorrefmark{3}
Woven by Toyota, Inc., Tokyo, Japan\\
quan.kong@woven.toyota}
}

\maketitle

\begin{abstract}
Event-camera--based visible light communication (VLC) is promising for communicating high-bandwidth data, since event cameras offer high temporal resolution with low data redundancy, which is suitable for high-speed LED modulation.
However, fixed-bit-bin demodulation based on event counts ignores temporal event spread across bit boundaries. Consequently, bit errors are affected by neighboring bit patterns.
In this work, we analyze the error characteristics of a single-transmitter event-camera VLC setup by focusing on a target bit and its neighboring bits. The results show that errors are concentrated in specific local polarity patterns, e.g., ``0101'' and ``0100''.
Based on this observation, we propose context-aware demodulation (CAD), which adjusts event weights and decision intervals based on the neighboring-bit context. We also propose an event-temporal convolutional neural network demodulation (ET-CNN) that learns the temporal distribution of events around each target bit. Indoor static and moving-transmitter experiments show that both methods improve demodulation performance. ET-CNN achieves a bit error rate (BER) below $10^{-4}$ and $192$~kbit/s at a communication distance of $3$~m.
\end{abstract}


\section{Introduction}

Visible light communication (VLC) transmits information by rapidly modulating light-emitting diodes (LEDs). Optical camera communication (OCC), in which a camera acts as the receiver, can receive communication signals while separating and locating the positions of different light sources~\cite{Saha2015OCCSurvey}.
However, the frame rate and exposure time of conventional cameras limit high-speed LED modulation and decoding.
Event cameras have recently emerged as a better-suited receiver for high-speed VLC.
They asynchronously detect brightness changes at each pixel, producing positive and negative ``events'' for brightness increases and decreases, respectively. Because they report only when and where changes occur, event cameras provide high temporal resolution with low data redundancy~\cite{Lichtsteiner2008DVS,Brandli2014DAVIS,Gallego2022pami,E-VLC-CVPRW-2025,soga}.

Several modulation schemes have been investigated for event-camera--based OCC, including on-off keying (OOK), pulse-position modulation (PPM), pulse-shaped waveform modulation, and schemes that encode information by the number of pulses~\cite{shen1,anrim,aranda2024npulse,Cubillo2025PulseShaped}. In OOK, LED on/off transitions generate positive and negative events. The receiver can therefore aggregate events within predefined time bins and make bit decisions from their polarities. This simple operation exploits the high temporal resolution of event cameras and provides a practical basis for high-throughput and robust communication.

\begin{figure}[!t]
\centering
\includegraphics[width=0.9\linewidth]{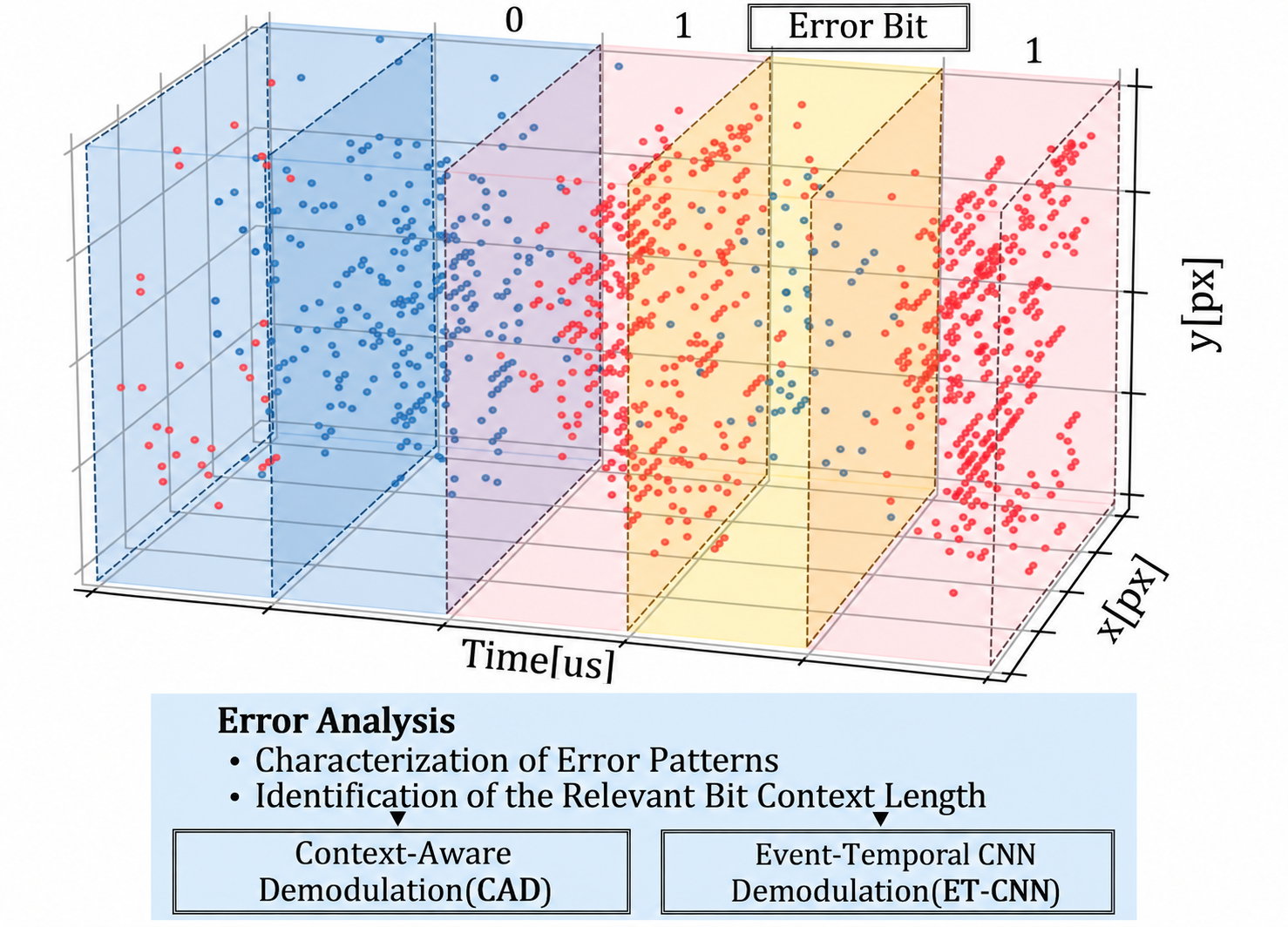}
\hspace{-2mm}
\caption{Motivation for context-aware event-camera VLC: temporal event spread can make neighboring-bit patterns affect fixed-bin bit decisions.}
\label{fig:intro_eyecatcher}
\end{figure}

Previous studies have investigated OOK-based event-camera VLC and proposed timing-correction methods.
Sumino \textit{et al.} propose an event-based OCC system combining non-return-to-zero OOK, toggle demodulation, and a digital phase-locked loop (DPLL)~\cite{sumino}.
Their DPLL corrects the clock phase from event peaks and suppresses burst errors caused by timing jitter and missed pulse detections, achieving BERs below $10^{-3}$ at $60$~kbit/s over $200$~m, $50$~kbit/s over $300$~m, and $20$~kbit/s over $400$~m.
Howard \textit{et al.} propose dynamic timing correction based on event timestamps to compensate for decision-timing offsets caused by scene motion and report an estimated channel capacity of approximately $75.8$~kbit/s~\cite{howard}.

These methods effectively compensate for clock-phase and bit-boundary errors; however, they do not account for variations in event timing and polarity within each bit interval. Events generated by an LED transition may spread into adjacent bit bins, causing neighboring bits to affect the target-bit decision. Moreover, the event distribution varies with the surrounding bit pattern, making bit-by-bit independent detection unreliable under certain conditions.

Our preliminary analysis further shows that event generation decreases immediately after polarity reversals and during runs of identical bits. In addition, neighboring-bit events may enter the target-bit interval and distort the polarity balance used for demodulation. These effects may be related to the temporal response or refractory behavior of event-camera pixels.

In summary, there are two challenges in existing event-camera VLC methods.
First, the influence of neighboring bit patterns on temporal event distributions and demodulation errors has not been quantitatively characterized.
Second, there is no demodulator that exploits the pattern dependence more effectively than fixed-bit-bin aggregation.

This work addresses these issues for VLC between a single LED transmitter and an event camera.
We first analyze pattern-dependent error characteristics. We then develop context-aware demodulation (CAD), based on this error analysis, and propose event-temporal CNN demodulation (ET-CNN) that effectively learns positive and negative event distributions around the target bit. 
Our principled analysis reveals bit-pattern-dependent error characteristics, which allows us to design an effective demodulation.
The experiments further demonstrate reliable high-rate communication at up to 192 kbit/s.

Our contributions are summarized as follows:
\begin{itemize}
    \item We quantitatively analyze the temporal distributions of event polarities, as well as the BER, for each target-bit and neighboring-bit pattern.
    \item We propose CAD, which adjusts event weights and decision intervals based on the neighboring-bit context.
    \item We propose ET-CNN, which leverages the temporal event distribution around the target bit, identified by the error analysis as contributing to bit errors.
    \item We evaluate both methods in indoor static and moving-transmitter experiments. At $3$~m and $137$~kbit/s, CAD reduces BER from $9.97\times10^{-3}$ with baseline demodulation to $2.95\times10^{-6}$. ET-CNN also outperforms baseline demodulation and achieves a BER below $10^{-4}$ at 3~m and $192$~kbit/s.
\end{itemize}


\section{Event-Camera-Based VLC System and Baseline Demodulation}
\label{sec:system}

\begin{figure*}[!t]
\centering
\includegraphics[width=0.73\textwidth]{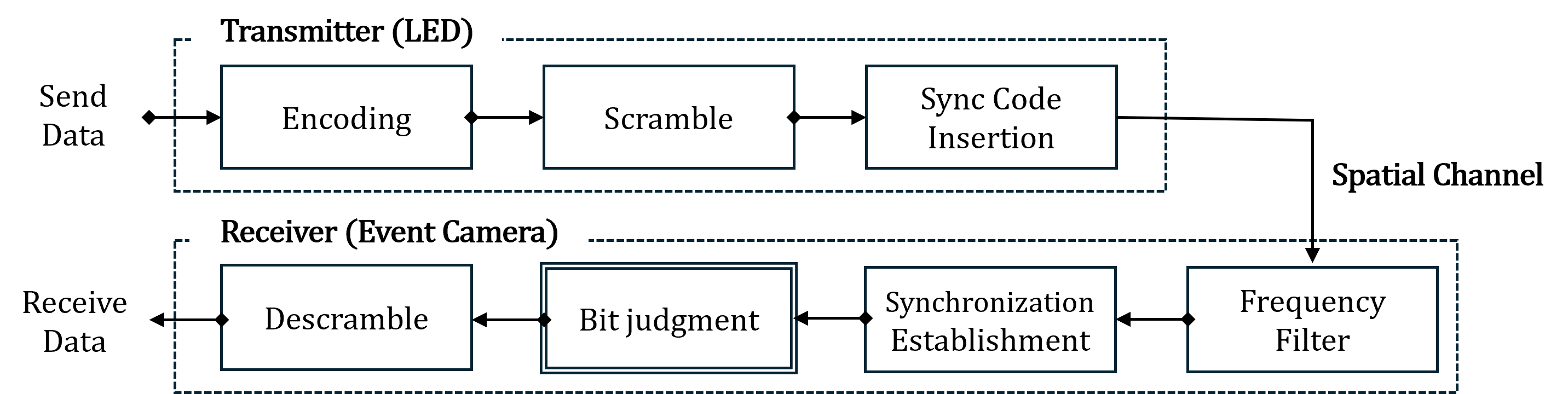}
\hspace{-2mm}
\caption{System model of the single-transmitter event-camera VLC system. The highlighted bit-judgment stage, which receives timing-aligned positive and negative events, is the focus of this work; baseline demodulation, CAD, and ET-CNN are compared at this stage.}
\label{fig:system}
\end{figure*}

This section describes the VLC system comprising an LED transmitter and an event-camera receiver. Figure~\ref{fig:system} illustrates the overall architecture of the event-camera-based VLC system. The transmitter uses OOK to send a packet consisting of a synchronization sequence, a scramble-ID field, and a scrambled-data field. The receiver captures LED brightness changes as positive and negative events, extracts the LED region, estimates timing, creates bit bins, and demodulates the received bits.

\subsection{Transmitted Packet Structure}

The transmitter maps LED on/off states to bit values using OOK. Each packet consists of a 26-bit synchronization sequence, a 15-bit scramble-ID field, and a 1024-bit scrambled-data field, for a total of 1065 bits. The synchronization sequence is used to estimate the packet start time and define the interval corresponding to each bit.

The scramble-ID field identifies the pseudo-random sequence used to scramble the data field. A 6-bit ID selects one of multiple pseudo-random sequences; the selected sequence is then XORed with the transmitted data. The receiver regenerates the selected sequence from the decoded ID and descrambles the received data. To improve ID recovery reliability, the 6-bit ID is encoded into a 15-bit BCH codeword.

Because an event camera responds to brightness changes, a long run of identical bits reduces event generation and degrades demodulation. Multiple scrambling candidates are therefore generated, and candidates containing runs of 10 or more identical bits in the concatenated scramble-ID and scrambled-data fields are excluded.

\subsection{Synchronization and Timing Estimation}

The receiver uses the 26-bit synchronization sequence to estimate the packet start time. It searches each pixel for an event-interval pattern corresponding to the known synchronization sequence. Pixels with dense matching events are used to estimate the initial timing and extract the LED signal region.

After initial synchronization, the receiver refines the timing using a phase-density distribution. Each event timestamp is folded modulo the bit period, and event density is calculated as a function of phase. Events are dense immediately after LED switching and sparse just before switching; thus, the phase of minimum density is selected as the bit boundary. 

\subsection{Bit Binning and Baseline Demodulation}

Using the corrected packet start time, the receiver defines a time interval for each bit, called a bit bin. Let $T_b$ be the bit period and $\hat{t}_0$ the corrected packet start time. The $i$-th bit bin is
\begin{equation}
\hat{t}_0+iT_b \leq t < \hat{t}_0+(i+1)T_b.
\end{equation}

For an event timestamp $t_k$, its bit index is
\begin{equation}
i_k=\left\lfloor\frac{t_k-\hat{t}_0}{T_b}\right\rfloor.
\end{equation}

We refer to this simple bit-bin method — which determines each bit from the difference between positive and negative event counts — as \emph{baseline demodulation}. Let $N_i^{+}$ and $N_i^{-}$ be the positive and negative event counts in the $i$-th bin. The decision value is
\begin{equation}
S_i=N_i^{+}-N_i^{-}.
\end{equation}
The estimated bit is
\begin{equation}
\hat{b}_i=
\begin{cases}
1, & S_i>0,\\
0, & S_i<0,\\
\hat{b}_{i-1}, & S_i=0.
\end{cases}
\end{equation}

Baseline demodulation is computationally lightweight, but it ignores temporal biases within a bit bin and event leakage from adjacent bits. It can therefore make errors when events span bit boundaries or when event generation depends on neighboring bit patterns.

\section{Error Analysis Based on Neighboring-Bit Patterns}
\label{sec:error_analysis}

\subsection{Error-Context Analysis}
\label{sec:error_context}

This section analyzes the contexts in which baseline demodulation errors occur. The event-camera biases were configured for high-speed LED modulation: \texttt{bias\_fo} (response speed) and \texttt{bias\_refr} (refractory period) were set to their maximum values of 55 and 235, respectively. Because these settings substantially increase negative-event noise, \texttt{bias\_diff\_off}, the negative-event contrast threshold, was increased to 15 for noise suppression. Although the quantitative results vary with the bias configuration, the same qualitative error tendencies were confirmed under different settings.

No event weighting or adaptive interval selection is applied. An error is recorded when the estimated bit $\hat{b}_i$ differs from the transmitted bit $b_i$. The analysis uses data bits in successfully synchronized packets for which the previous three and next two bits are available.

To determine the extent of neighboring context needed to explain errors, we compare six context models. M0 uses $(b_i)$; M1 uses $(b_{i-1},b_i)$; M2 uses $(b_{i-2},b_{i-1},b_i)$; M3 uses $(b_{i-2},b_{i-1},b_i,b_{i+1})$; M4 uses $(b_{i-3},b_{i-2},b_{i-1},b_i,b_{i+1})$; and M5 uses $(b_{i-2},b_{i-1},b_i,b_{i+1},b_{i+2})$. For each context, we estimate the conditional error probability and calculate LogLoss on the evaluation data. The improvement over M0 is evaluated by the reduction in LogLoss, where a larger reduction indicates that the additional context better explains error locations.

\begin{figure}[t]
\centering
\includegraphics[width=0.9\linewidth]{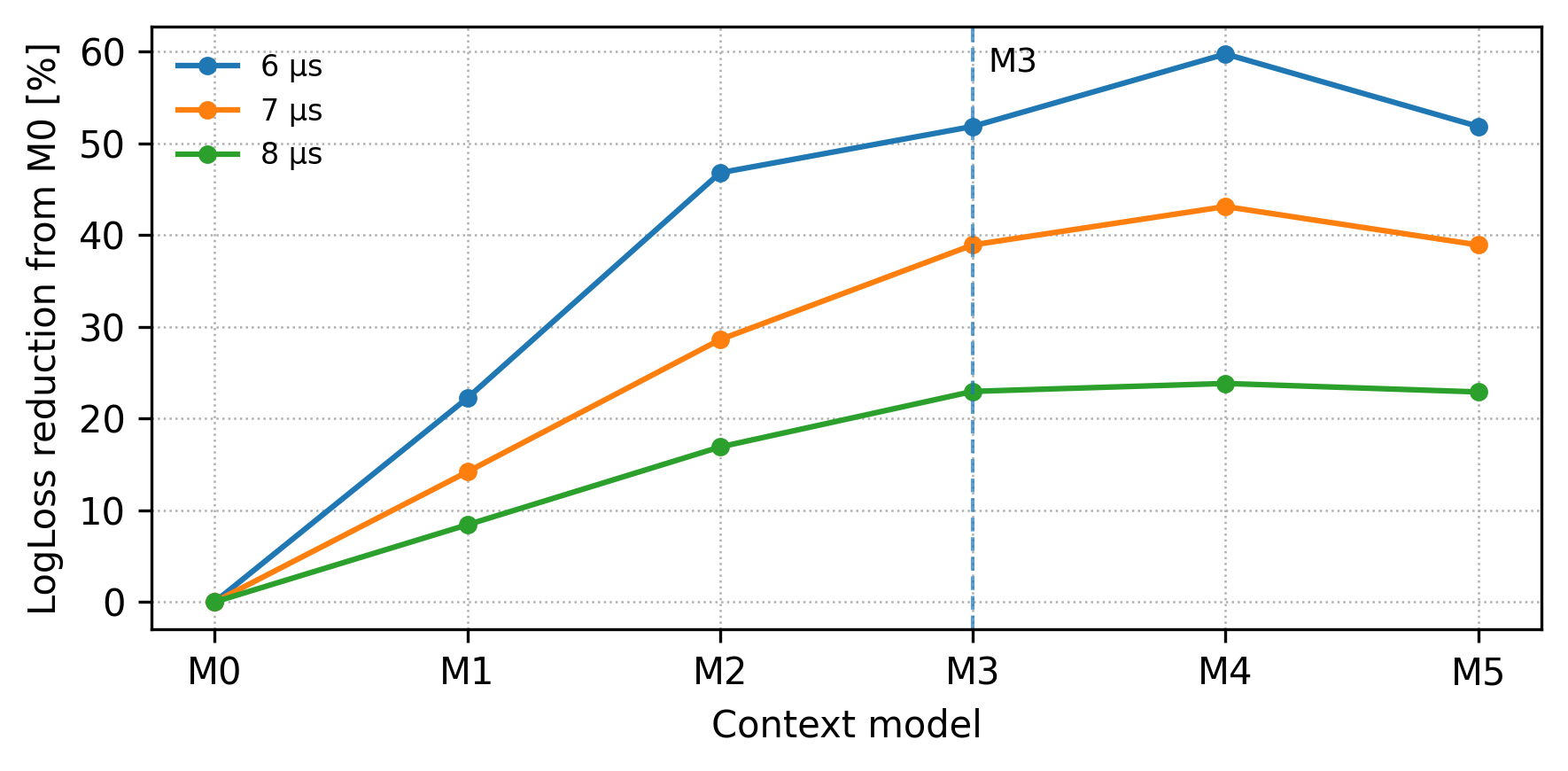}
\hspace{-4mm}
\caption{LogLoss reduction from M0 for the error-context models at bit periods of 6, 7, and 8~$\mu\mathrm{s}$. Higher values indicate a greater improvement in explaining baseline-demodulation errors.}
\hspace{-2mm}
\label{fig:context_logloss}
\end{figure}

Fig.~\ref{fig:context_logloss} shows the LogLoss reduction from M0 for $T_b=6$, 7, and 8~$\mu\mathrm{s}$.

Extending the context from M0 to M3 increases the LogLoss reduction for every evaluated bit period, indicating that the two preceding bits and the following bit are important for explaining error locations. Although M4 provides an additional improvement, M3 captures most of the context dependence with a shorter context and is therefore used in the following pattern analysis. Thus, baseline-demodulation errors depend strongly on local polarity patterns.

We therefore examine how errors are distributed over every four-bit context $(b_{i-2},b_{i-1},b_i,b_{i+1})$. Table~\ref{tab:error_context_summary} shows the fractions of all errors occurring in the dominant contexts 0101 and 0100.

\begin{table}[t]
\centering
\caption{Fractions of baseline-demodulation errors occurring in the dominant four-bit contexts 0101 and 0100.}
\label{tab:error_context_summary}
\begin{tabular}{c|c|c}
\hline
$T_b$ [$\mu$s] & 0101 & 0100 \\
\hline
6 & 72.5\% & 27.3\% \\
7 & 94.2\% & 5.8\% \\
8 & 95.4\% & 4.6\% \\
\hline
\end{tabular}
\end{table}

Errors are strongly concentrated in these two patterns rather than uniformly distributed across the 16 contexts. In particular, 0101 accounts for 72.5\%, 94.2\%, and 95.4\% of the errors at $T_b=6$, 7, and 8~$\mu\mathrm{s}$, respectively. Since 0101 and 0100 share the preceding sequence 010, their event distributions are analyzed next to identify the cause of this pattern-dependent behavior.

\subsection{Event-Count Analysis of Error Causes}
\label{sec:error_cause}

Because errors are concentrated in 0101 and 0100, we analyze their temporal event-count distributions together with the polarity-inverted contexts 1010 and 1011.

For all four contexts, the number of events with the polarity expected in the third bit is lower than that in control contexts without a polarity reversal between the first and second bits, decreasing to as little as approximately 0.2 times the control level. In addition, during the first and last 2~$\mu\mathrm{s}$ of the third bit, opposite-polarity events attributable to neighboring bits outnumber the expected-polarity events. These effects become more pronounced at shorter bit periods.

These results identify two main sources of error. First, a polarity reversal between the first and second bits suppresses events with the expected polarity in the third bit. Second, opposite-polarity events from neighboring bits enter the decision interval of the target bit. The next section proposes demodulation methods to address these two causes.

\section{Demodulation Methods Based on Error Characteristics}
\label{sec:demod_methods}

\subsection{Context-Aware Demodulation}

The analyses in Sections~\ref{sec:error_context} and~\ref{sec:error_cause} show that specific contexts cause errors by suppressing expected-polarity events and contaminating them with events from neighboring bits. We therefore propose context-aware demodulation (CAD), which adjusts event weighting and decision intervals bit-by-bit. Context-based weighting emphasizes expected-polarity events to compensate for their decline, while adaptive interval selection avoids time regions where neighboring-bit events are likely to contaminate the decision.

\subsubsection{Weighted Event-Sum Decision}

CAD assigns a weight $w_i$ to each event and decides each bit from the weighted polarity sum. Let $\mathcal{E}_k$ be the event set used to decide bit $k$, and let $p_i\in\{+1,-1\}$ be the polarity of event $i$. The decision value is
\begin{equation}
S_k=\sum_{i\in\mathcal{E}_k}w_i p_i.
\end{equation}
The weight is determined by the event polarity, its timing within the bit, and previously decoded bits. CAD decides 1 for $S_k>0$, 0 for $S_k<0$, and retains the previous decision for $S_k=0$.

\subsubsection{Context-Based Event Weighting}

As shown in Section~\ref{sec:error_cause}, the expected-polarity event count in the third bit decreases after a polarity reversal. To compensate for this decrease, CAD determines the polarity to emphasize from the two previously decoded bits. When the previous bits are 10, positive events are emphasized; when they are 01, negative events are emphasized. During runs of identical bits, the corresponding-polarity event count also decreases from the second bit onward; events with that polarity therefore receive additional weight according to the run length. Events near the end of a bit may originate from the subsequent bit and are less reliable for the current decision; CAD therefore suppresses their additional weighting. This weighting compensates for decreased expected-polarity events while reducing errors caused by event leakage from neighboring bits.

\subsubsection{Adaptive Decision Interval}

CAD selects the decision interval in two steps. First, it searches for the decision start time near the fixed boundary. Let
\begin{equation}
t_k^{\mathrm{fix}}=\hat{t}_0+kT
\end{equation}
be the fixed start time of bit $k$, where $\hat{t}_0$ is the estimated packet start time and $T$ is the bit period. The search range is
\begin{equation}
[t_k^{\mathrm{fix}}+\Delta_{\min},\,t_k^{\mathrm{fix}}+\Delta_{\max}].
\end{equation}
As shown in Section~\ref{sec:error_cause}, opposite-polarity events from neighboring bits are likely to enter the beginning and end of the third-bit interval. CAD, therefore, focuses on polarity reversals in the error-prone contexts 0100 and 0101: it searches for positive events as transition candidates after a decoded 0 and for negative events after a decoded 1. The earliest candidate is accepted when events of the candidate polarity occur continuously near it, and the weighted polarity sum in a short window agrees with that polarity. If no candidate satisfies these conditions, the fixed start time is used.

Second, CAD examines a tail interval immediately before the fixed end time,
\begin{equation}
[t_k^{\mathrm{fix,end}}-\delta,\,t_k^{\mathrm{fix,end}}),
\end{equation}
where $t_k^{\mathrm{fix,end}}=t_k^{\mathrm{fix}}+T$. The tail interval is excluded if its polarity matches that of the previous bit and its removal would change the resulting bit decision. The weighted decision is then calculated over
\begin{equation}
S_k=\sum_{i:\,t_k^{\mathrm{start}}\leq t_i<t_k^{\mathrm{end}}}w_i p_i.
\end{equation}
\enlargethispage{-1.3mm}
Even after event weighting and interval selection, negative events may not be sufficiently dominant in the third bit of 0100 and 0101. Therefore, after the first decoding pass, a preceding bit with an extremely small decision value and observed negative events is treated as an error candidate and inverted. The inverted bit sequence is used as context, and demodulation is repeated once.

CAD thus compensates for the decrease in expected-polarity events after a polarity reversal through weighting, suppresses contamination by neighboring-bit events through interval selection, and corrects remaining error candidates through one re-demodulation pass. Algorithm~\ref{alg:cad} summarizes this procedure. In each decoding pass, only previously decoded bits are used as the context of the current bit.

\begin{algorithm}[t]
\caption{CAD procedure with context-aware interval selection, event weighting, and one re-demodulation pass.}
\label{alg:cad}

\begin{algorithmic}[1]

\REQUIRE Corrected start time $\hat{t}_0$, bit period $T$, and event stream $\{(t_i,p_i)\}$
\ENSURE Decoded bits $\{\hat{b}_k\}$

\FOR{each bit $k$}
\STATE Obtain the preceding decoded bits $(\hat{b}_{k-2},\hat{b}_{k-1})$
\STATE Determine event weights $w_i$ from polarity and run length
\STATE Search near $t_k^{\mathrm{fix}}=\hat{t}_0+kT$ for a transition candidate based on $\hat{b}_{k-1}$
\STATE Set $t_k^{\mathrm{start}}$ to the accepted candidate, or to $t_k^{\mathrm{fix}}$ if none is found
\STATE Set $t_k^{\mathrm{end}}$ while excluding a tail region likely affected by the next bit
\STATE Calculate
\[
S_k=\sum_{i:\,t_k^{\mathrm{start}}\le t_i<t_k^{\mathrm{end}}} w_i p_i
\]
\STATE Decide $\hat{b}_k=1$ if $S_k>0$, $\hat{b}_k=0$ if $S_k<0$, and $\hat{b}_{k-1}$ otherwise
\ENDFOR

\STATE Identify likely preceding-bit errors from small decision values and negative events, invert them, and repeat demodulation once

\end{algorithmic}
\end{algorithm}

\subsection{Event-Temporal CNN Demodulation}

CAD uses manually designed weights and interval-selection rules. Because event distributions are also affected by pixel response, noise, and synchronization errors, designing optimal rules for all conditions is difficult. We therefore construct event-temporal CNN demodulation (ET-CNN), which learns the temporal event distribution around each target bit using a one-dimensional convolutional neural network.

\subsubsection{Input Features and Network}

Let the corrected start time of bit $k$ be
\begin{equation}
t_k=\hat{t}_0+kT.
\end{equation}
For each target bit, a time window including the target and neighboring bits is divided into 1-$\mu$s subbins. For every subbin, the positive count $P$, negative count $N$, and polarity ratio
\begin{equation}
R=\frac{P-N}{P+N+\epsilon}
\end{equation}
are calculated, where $\epsilon$ avoids division by zero. The input is a three-channel temporal sequence of $P$, $N$, and $R$.

\begin{figure}[t]
\centering
\includegraphics[width=0.83\linewidth]{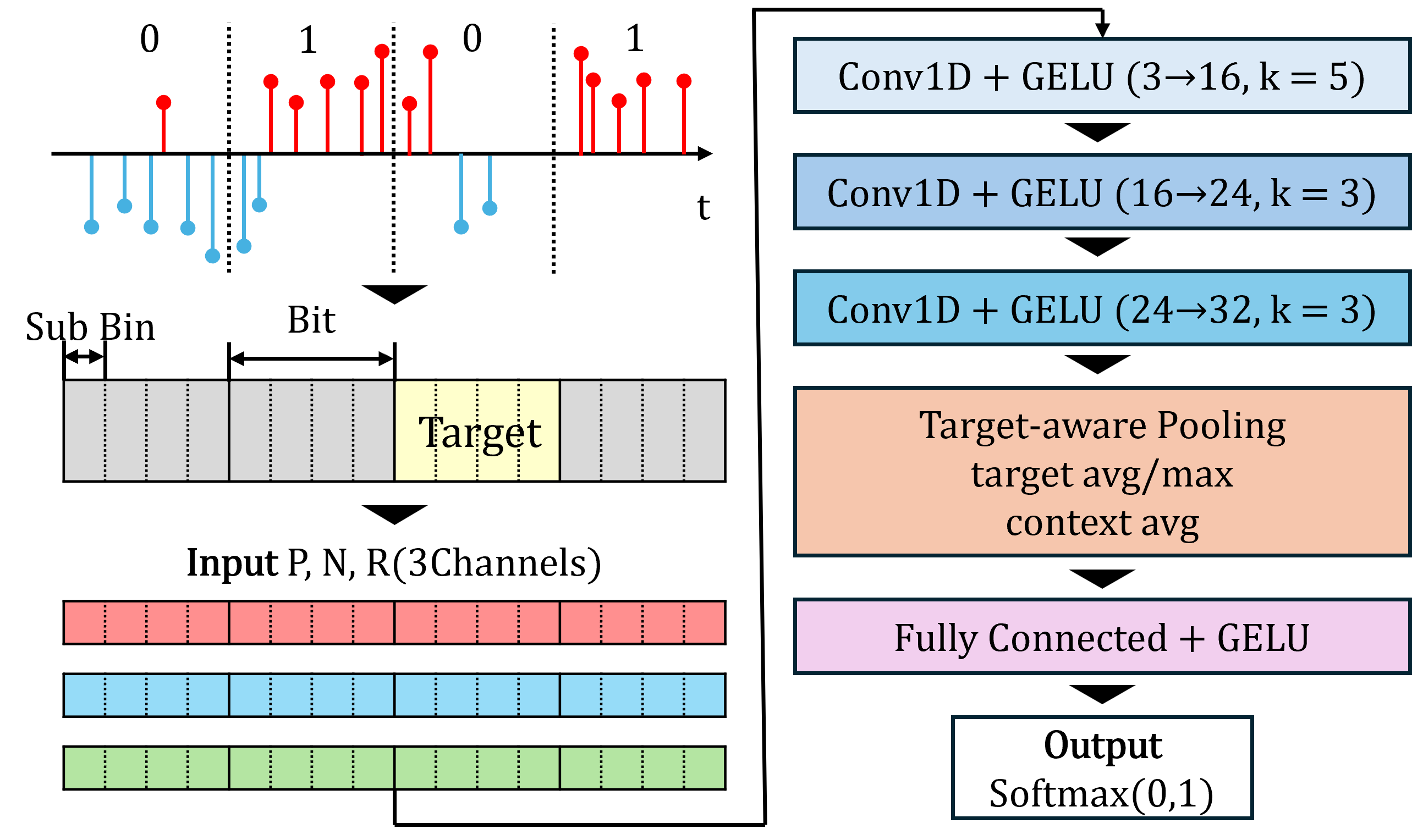}
\caption{Architecture of ET-CNN. Positive-event count $P$, negative-event count $N$, and polarity ratio $R$ form a three-channel temporal input; convolutional and pooling features from the target and neighboring bits are classified as $\hat{b}_k$.}
\label{fig:etcnn_architecture}
\end{figure}

The network applies one-dimensional convolutions to extract local temporal patterns. Average and maximum features are extracted from the target-bit region, and average features are extracted from the preceding and following contexts. These features are concatenated and passed to a fully connected layer that classifies the target bit as 0 or 1. Fig.~\ref{fig:etcnn_architecture} shows the ET-CNN architecture.

\subsubsection{Training}

Each data bit in a valid packet is treated as one training sample. Labels are reconstructed from the transmitted sequence using the decoded random-sequence identifier rather than using the demodulated bits. Training, validation, and test sets are split by packet to prevent bits from the same packet from appearing in multiple sets. Positive and negative counts are log-transformed, and all input channels are standardized. The network is trained with cross-entropy loss and early stopping based on validation loss.

\section{Experimental Evaluation in Indoor Environments}
\label{sec:evaluation}

\subsection{Indoor Static Experiment}

\begin{figure}[t]
\centering
\includegraphics[width=0.88\linewidth]{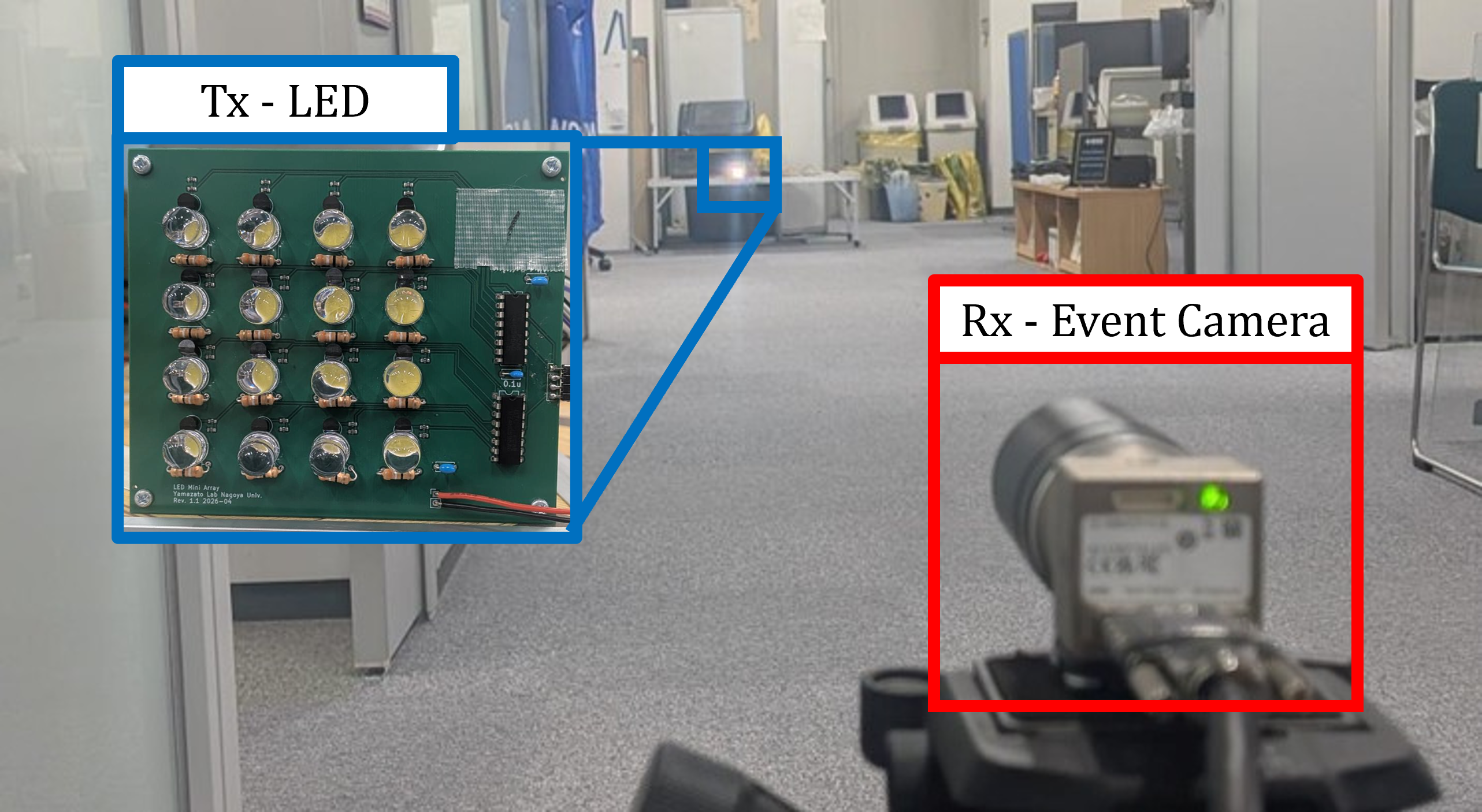}
\caption{Indoor static-transmitter experimental setup using four OS4WMLA131A LEDs and an IMX636 event camera at distances of 3, 4, and 5~m.}
\label{fig:indoor_static_setup}
\end{figure}

The transmitter and receiver were fixed in an indoor environment at night with an illuminance of approximately 550~lx. The transmitter used four of the 16 OS4WMLA131A LEDs mounted on the transmitter board, driven by a Raspberry Pi Pico 2. The receiver was an IDS UE-39B0XCP-E event camera with an IMX636 sensor, a resolution of $1280\times720$ pixels, and a 4-mm lens. The \texttt{refr} and \texttt{fo} bias parameters, which control event-generation sensitivity and temporal response, were set to their maximum values. Transmission distances were 3, 4, and 5~m. Bit periods of 5, 6, 7, 8, and 9~$\mu\mathrm{s}$ correspond to transmission rates of 192, 160, 137, 120, and 106~kbit/s, respectively. For each condition, 5000 packets were transmitted. Fig.~\ref{fig:indoor_static_setup} shows the experimental setup.

We compare baseline method, CAD, and ET-CNN. ET-CNN was trained using data separately acquired at 3 and 5~m, while CAD parameters were empirically selected using separate data. Random sequences were transmitted in the data field, and the receiver reconstructed the transmitted data from the decoded sequence identifier. Packets for which synchronization, the scramble ID, and the random-sequence identifier were successfully recovered were treated as valid. BER was calculated from the 1024-bit data field of valid packets.

\begin{figure}[!t]
\centering
\includegraphics[width=0.9\linewidth]{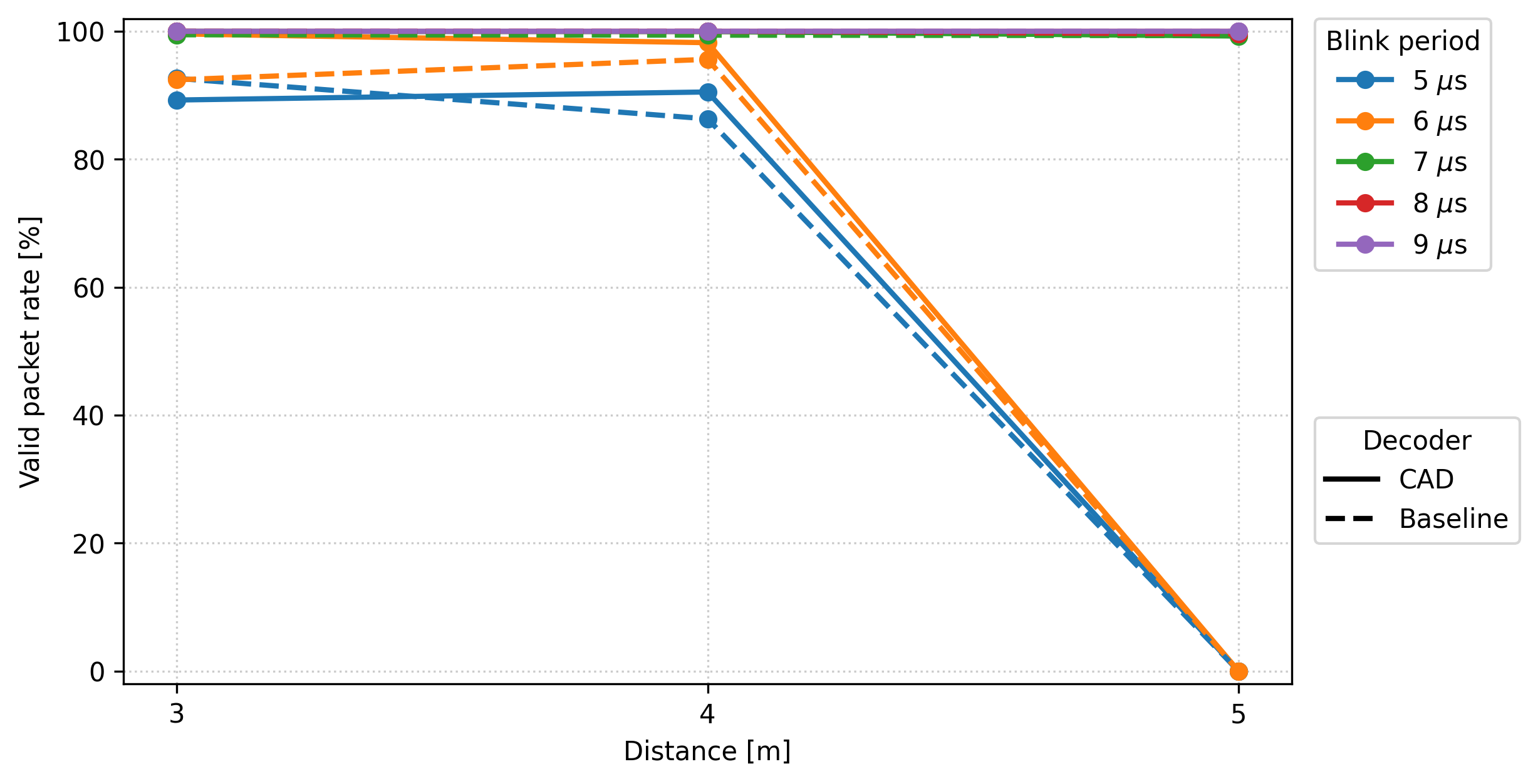}\par
{\footnotesize (a) Valid-packet rate versus communication distance at bit periods of 5--9~$\mu\mathrm{s}$.}\par\medskip
\includegraphics[width=0.9\linewidth]{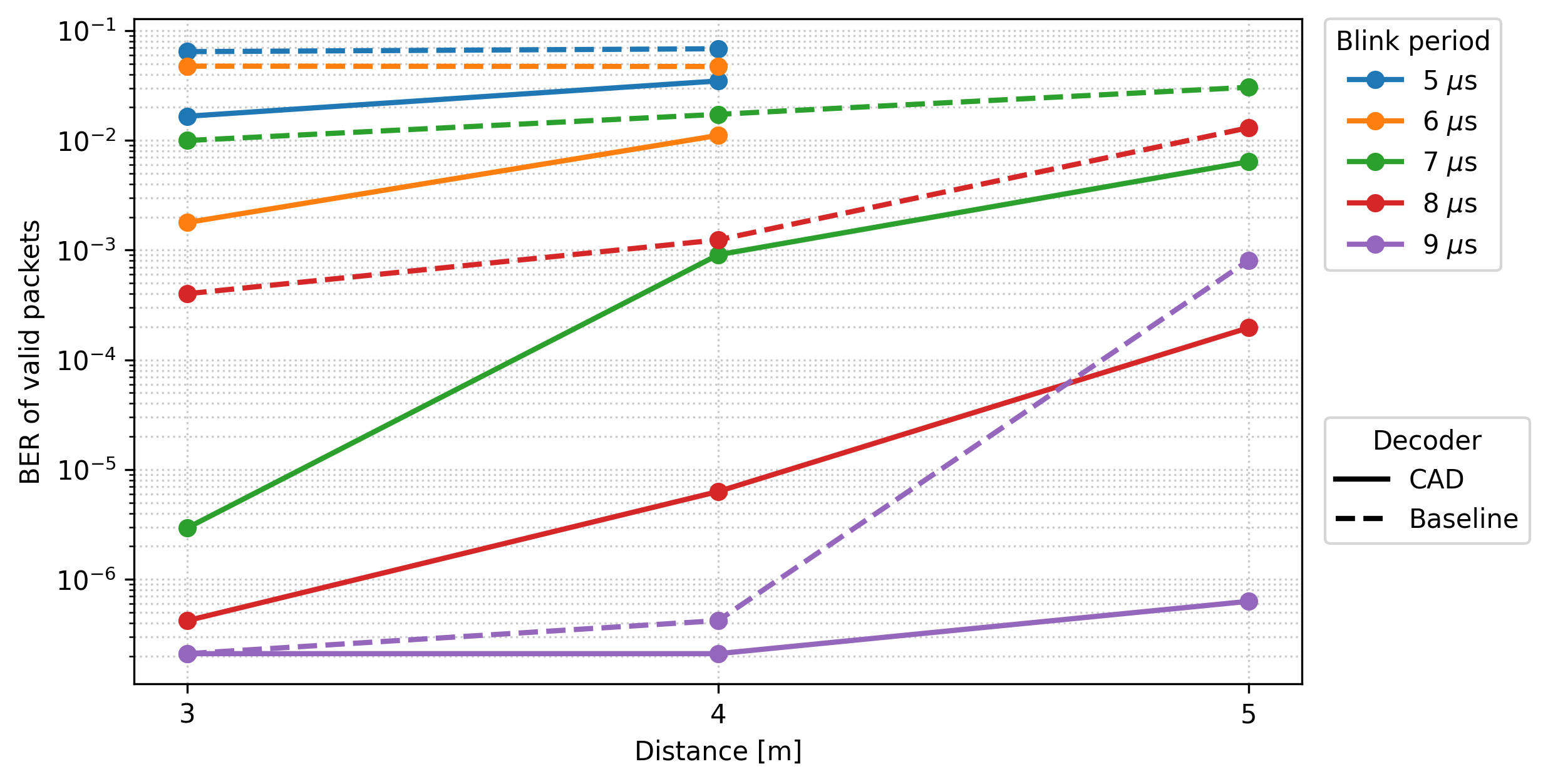}\par
{\footnotesize (b) BER of valid packets versus communication distance. BER is calculated from the 1024-bit data field of valid packets.}
\hspace{-2mm}
\caption{Performance comparison between baseline demodulation and CAD in the indoor static experiment.}
\label{fig:indoor_static_baseline_cad}
\end{figure}

\begin{figure}[!t]
\centering
\includegraphics[width=0.9\linewidth]{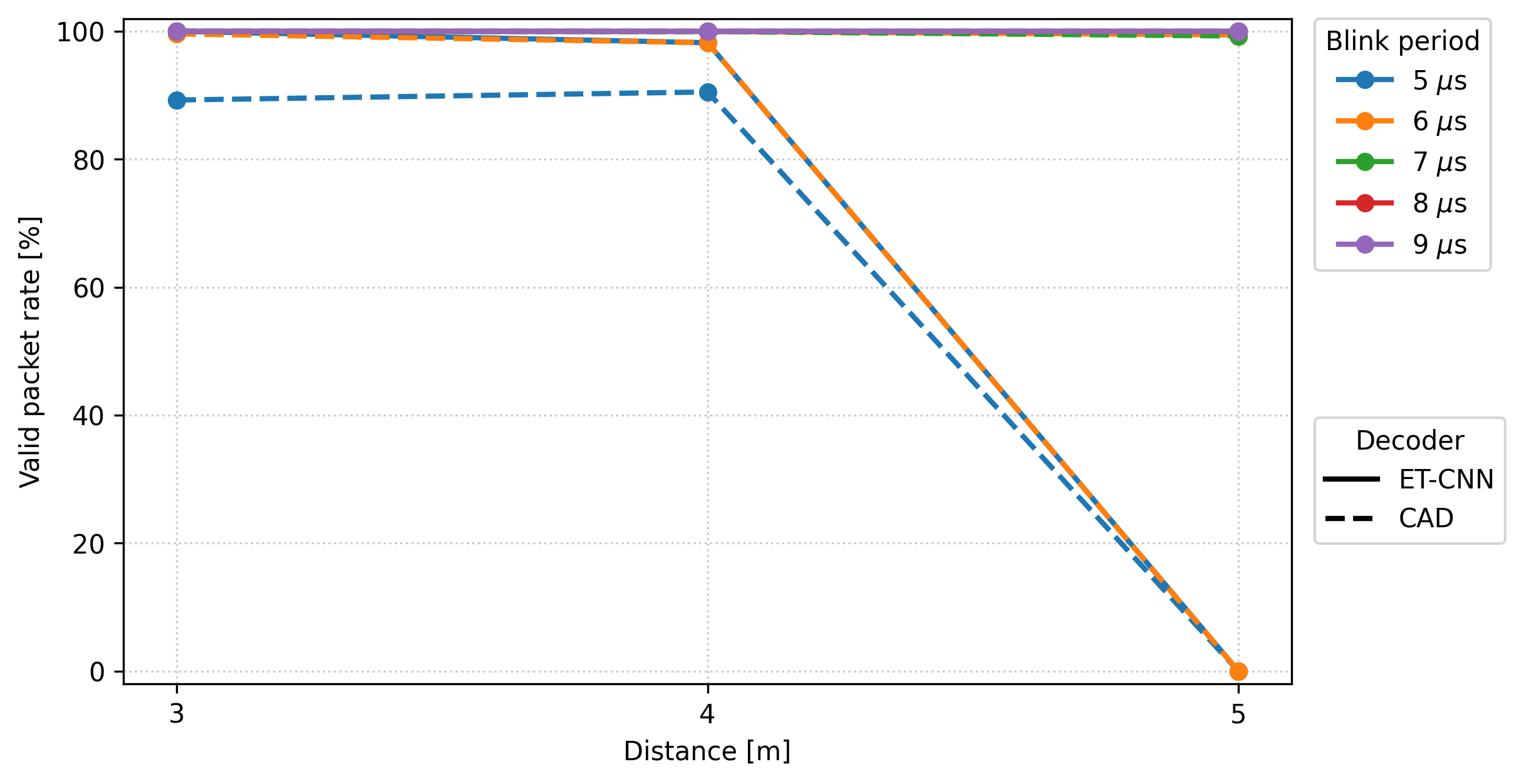}\par
{\footnotesize (a) Valid-packet rate versus communication distance. Each point is based on 5000 transmitted packets; colors indicate bit periods of 5--9~$\mu\mathrm{s}$ (192--106~kbit/s).}\par\medskip
\includegraphics[width=0.9\linewidth]{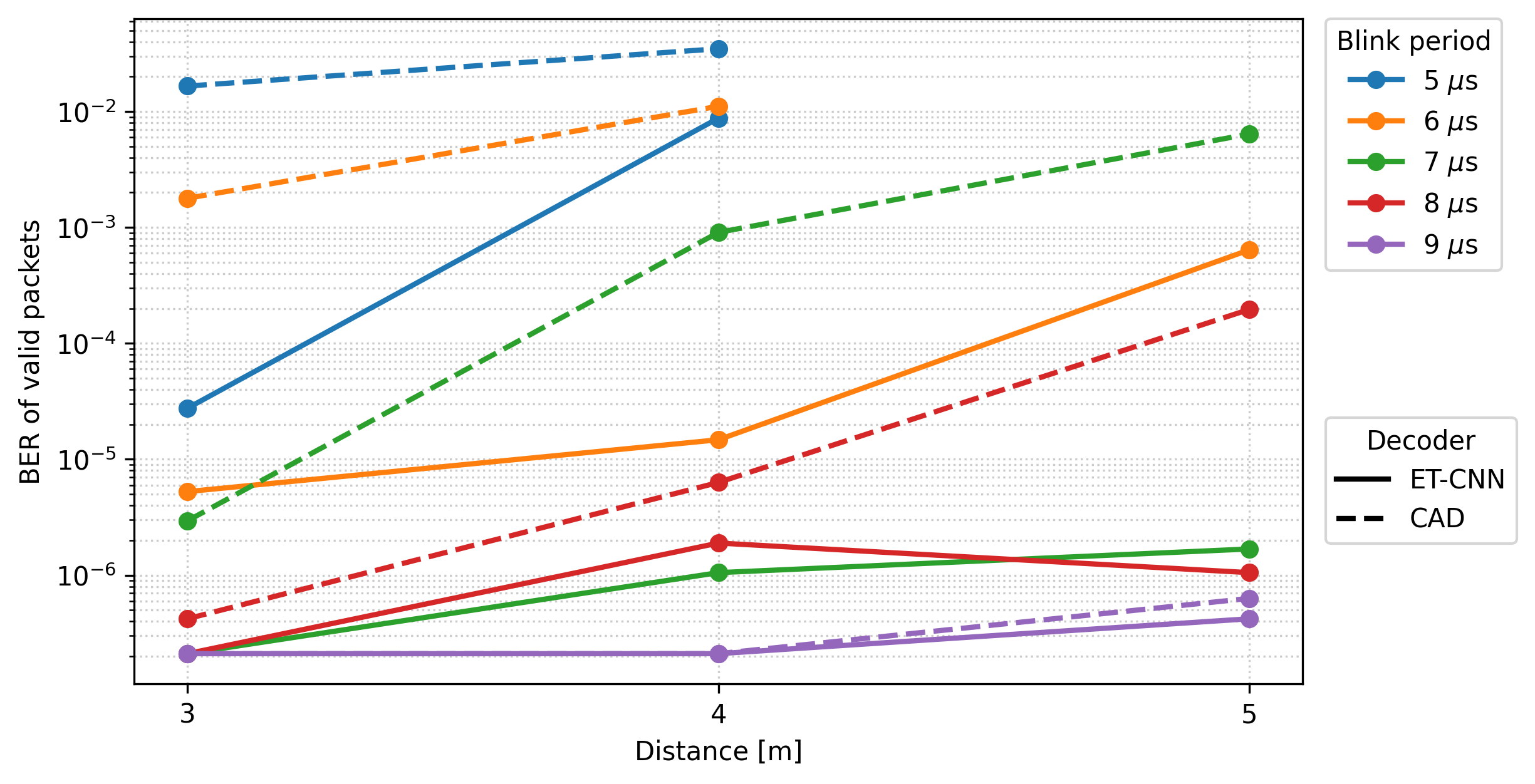}\par
{\footnotesize (b) BER of valid packets versus communication distance. Colors indicate bit periods of 5--9~$\mu\mathrm{s}$ (192--106~kbit/s).}
\caption{Performance comparison between CAD and ET-CNN in the indoor static experiment.}
\label{fig:indoor_static_cad_etcnn}
\end{figure}

Figures~\ref{fig:indoor_static_baseline_cad} and \ref{fig:indoor_static_cad_etcnn} show the results. At short bit periods, the valid-packet rate decreases with communication distance. ET-CNN maintains a higher valid-packet rate than CAD and baseline demodulation. Both proposed methods reduce BER relative to baseline demodulation. In particular, at 3~m and a 7-$\mu\mathrm{s}$ bit period (137~kbit/s), CAD reduces BER from $9.97\times10^{-3}$ with baseline demodulation to $2.95\times10^{-6}$. ET-CNN achieves the lowest BER among the evaluated methods and a BER below $10^{-4}$ at 3~m and 192~kbit/s.

\subsection{Indoor Moving-Transmitter Experiment}

The transmitter was moved in a reciprocating motion over 7.5~cm every 0.72~s, corresponding to approximately 10.4~cm/s, while the communication distance was fixed at 3~m. Bit periods from 5 to 9~$\mu\mathrm{s}$ were evaluated and compared with the static condition. CAD and ET-CNN were evaluated.

\begin{figure}[t]
\centering
\includegraphics[width=0.9\linewidth]{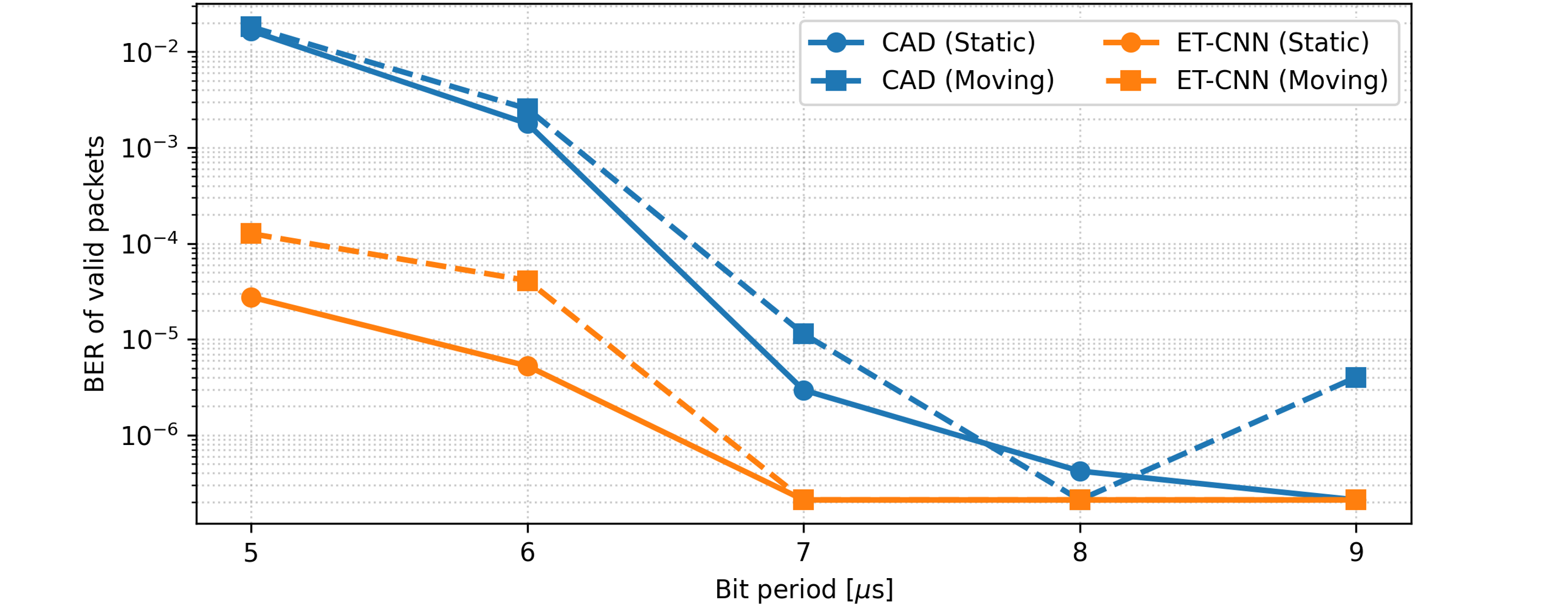}
\caption{BER comparison at 3~m between static and moving transmitter conditions: (a) CAD and (b) ET-CNN. The transmitter moves at approximately 10.4~cm/s; bit periods range from 5 to 9~$\mu\mathrm{s}$.}
\hspace{-2mm}
\label{fig:indoor_static_moving_ber}
\end{figure}

Fig.~\ref{fig:indoor_static_moving_ber} shows the BER. CAD exhibits a similar BER tendency under static and moving conditions, although its BER increases in some conditions. ET-CNN shows a modest BER increase at the 5- and 6-$\mu\mathrm{s}$ bit periods under motion, whereas its BER at 7--9~$\mu\mathrm{s}$ remains comparable to the static condition. ET-CNN successfully receives 5000 valid packets under every evaluated condition.

\section{Conclusion}
\label{sec:conclusion}

This paper analyzed the error characteristics of event-camera-based VLC by focusing on local bit patterns around a target bit. Baseline-demodulation errors were concentrated in 0101 and 0100. Based on this observation, we proposed CAD, which adapts event weighting and decision intervals to the neighboring-bit context, and constructed ET-CNN, which learns temporal event distributions.

Indoor static experiments showed that both proposed methods reduce BER relative to baseline demodulation. At 3~m and 137~kbit/s, CAD reduces BER from $9.97\times10^{-3}$ to $2.95\times10^{-6}$. ET-CNN achieves the lowest BER among the evaluated methods, reaching $2.76\times10^{-5}$ at 3~m and 192~kbit/s. It also achieves BERs of $1.90\times10^{-6}$ or lower for bit periods from 7 to 9~$\mu\mathrm{s}$ across all evaluated distances. In indoor moving-transmitter experiments at 10.4 cm/s, ET-CNN successfully received all 5000 packets under all conditions. Its BER increased only at the shortest evaluated bit periods and remained comparable to the static case at 7--9~$\mu\mathrm{s}$.

Future work includes evaluations over a wider range of transmitter velocities and communication distances, robustness evaluations under illumination variation and background noise, automatic optimization of CAD parameters, and lightweight ET-CNN models for real-time demodulation.

\bibliographystyle{IEEEtran}
\bibliography{soga}

\end{document}